\documentclass[12pt]{article}
\usepackage[utf8]{inputenc}
\usepackage{graphicx}
\usepackage{cite}
\usepackage{enumerate}

\providecommand{\keywords}[1]
{
  \small    
  \textbf{\textit{Keywords---}} #1
}

\begin{document}

  \title{Autocalibration Subsystem for Cable-Driven Parallel Robots}
  \author{Jaâfar Moussaid}
  \date{December {$19^{th}$}, 2019}
  \maketitle
  \thispagestyle{empty}

\begin{abstract}
For Cable-Driven Parallel Robots, the cable length is the only element we control. Better knowledge of this parameter will improve robot performances. We propose a new cable length \emph{autocalibration subsystem}. It consists of an \emph{instrumented cable}, \emph{inductive sensors} and a \emph{controller} for digital data processing. Cable would be instrumented by metallic marks, with variation in distance between successive marks. Inductive sensors will allow the detection of metallic marks; it will be placed vertically on the support, with variation in distance between successive sensors. Each detection will correspond to a cable length. By analyzing the succession of cable lengths detected at particular moments, we can identify our position on the cable, then deduce its length. A closed-loop autocalibration may allow continuous cable length adjustment.
\end{abstract}
\keywords{Cable-Driven Parallel Robot, Auto-calibration, Subsystem, Sensor, Encoder, Controller, Length, Measurement, Adjustment}

\section{Introduction}\label{sec:int}

One of levers to enhance performance is to question measure. Length of cable is the main parameter in determining state of a \emph{Cable-Driven Parallel Robot} (CDPR)~\cite{merlet2000direct}~\cite{pott2018cable}. The use of a motorized winch equipped with a rotary encoder, to control cable length, can be suitable for small or medium-sized CDPRs but poses a problem for large ones. The cable winding on the drum is generally done in two ways :
\begin{enumerate}[(i)]
  \item the drum has a spiral-guided mechanism moves cable synchronously over the free part~\cite{pott2013ipanema}, CoGiRo~\cite{gouttefarde2012simplified};
  \item cable is simply wound on the drum~\cite{merlet2010marionet}.
\end{enumerate}\
Measurements by an encoder, based on the winch drum rotation multiplying by its radius are \emph{incremental}\footnote{Measurement regarding the lastest movement made.}. This type of measurement may not be reliable, due to the ignorance of the drum radius. It is, therefore, necessary, when CDPR is launched, to estimate initial cable length, with two successive displacements and calculate mobile platform pose~\cite{dit2014certified},~\cite{chen2013integrated},~\cite{baczynski2010simple},~\cite{miermeister2012auto}. With this method the rotary encoder may provide an accurate measurement of cable length, but has inconveniences:
\begin{itemize}
  \item the drum may only accept one layer, such a mechanism may not be suitable for very large CDPRs, which involve winding of tens of cable meters;
  \item cable friction on spiral-guide\footnote{Spiral-guide distribute the cable well on the drum.} mechanism increases cable wear;
  \item depending on cable tension, the cable may jump to another part of the spiral-guide.
\end{itemize}

At start-up, calibration makes it possible to define the \emph{zero}\footnote{Corresponds to starting position.} of cables length; in operation, it consists of \emph{comparing} rotary encoder measurements with those of a standard calibration of known accuracy. Measurement accuracy may be different from the calibration one. The comparison result is one of the following cases:
\begin{enumerate}
  \item no error noted on cable length;
  \item error noted, no adjustment made;
  \item adjustment made, to correct error to an acceptable level.
\end{enumerate}
We distinguish two types of calibration:
\begin{enumerate}
  \item Manual calibration: operator manually enters corresponding cable length;
  \item Automatic calibration: CDPR performs calibration independently.
\end{enumerate}

We propose an automatic calibration method (auto-calibration), to adjust cable length at particular moments. Using metallic marks, placed on the cable, combined with inductive sensors on CDPR support. We demonstrate that this disposition permits automating calibration problem, using scheduling theory~\cite{framinan2014manufacturing},~\cite{glaschenko2009multi}.

\begin{figure}[!h]
  \centering
  \includegraphics[width=.8\linewidth]{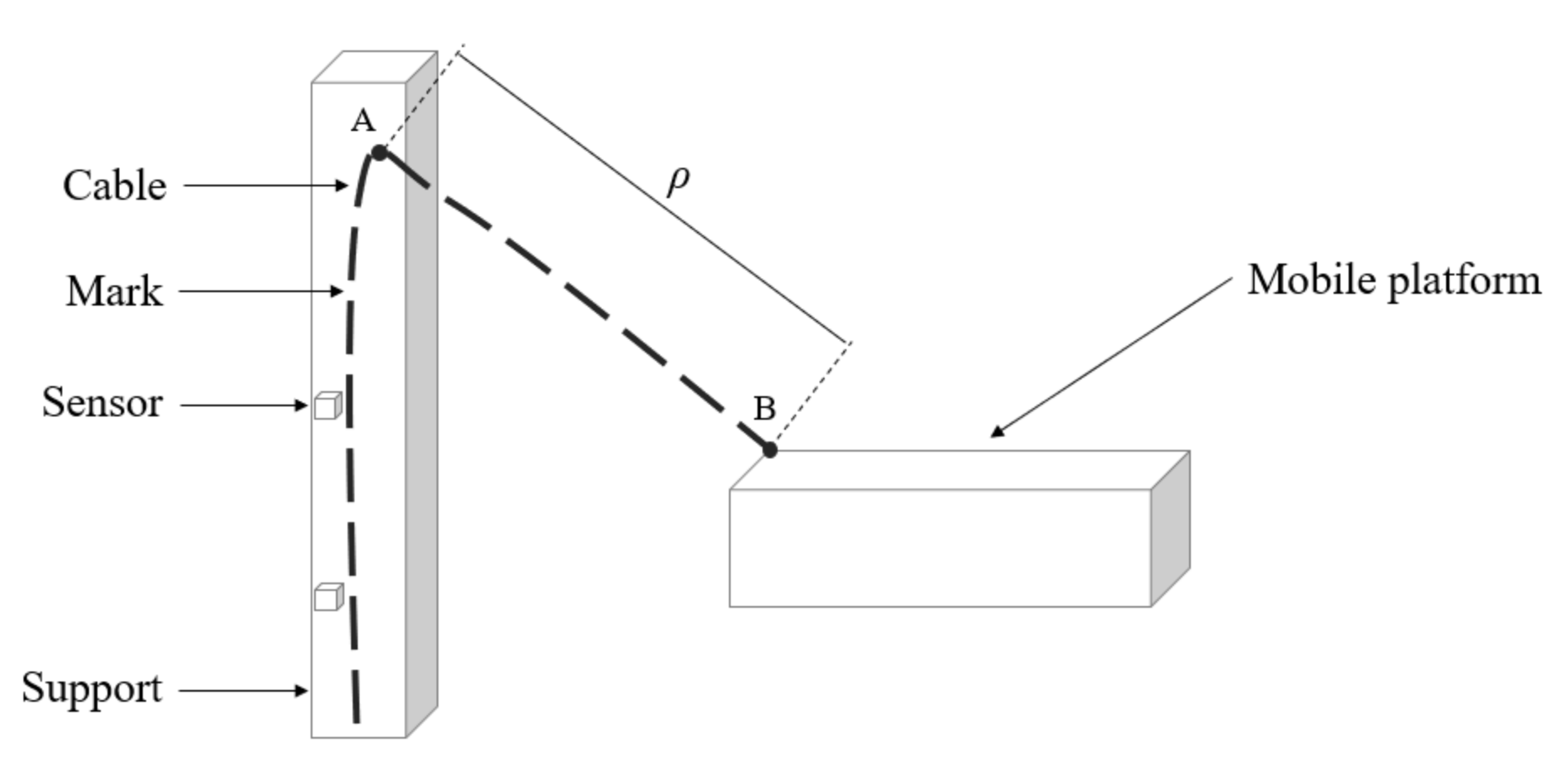}
  \caption{Instrumented cable principle}
  \label{fig:f1}
\end{figure}

Our method is inspired by a method successfully implemented on Marionet-Assist CDPR~\cite{merlet2019improving}. This CDPR uses synthetic cables and several marks of different colours\footnote{Three types of colors: red, green or blue.}, fixed on the cable at a constant distance at the beginning which increases towards the end, from proximal point A on CDPR support to distal point B on the mobile platform. At CDPR support, the cable passes through color sensors. Each time a color mark passes through a color sensor, an electrical signal is established, providing boolean information to the controller. Using this event, a semi-automatic procedure can be used for initialisation\footnote{Robot stops when a mark is detected and the operator manually enters corresponding cable length.} and, in operation, to update estimated cable length. We propose to improve this method, by using only one type of mark (\emph{metallic mark}\footnote{For example, aluminum foils.}) (Figure~\ref{fig:f1}).

\emph{Inductive sensors} on CDPR support make it possible to detect metallic mark passage. The distance variation between successive marks on cable, combined with distance variation between successive inductive sensors, will allow \emph{cables length auto-calibration}.
%\clearpage
\section{Instrumented cable}

\begin{figure}[!h]
  \centering
  \includegraphics[width=1\linewidth]{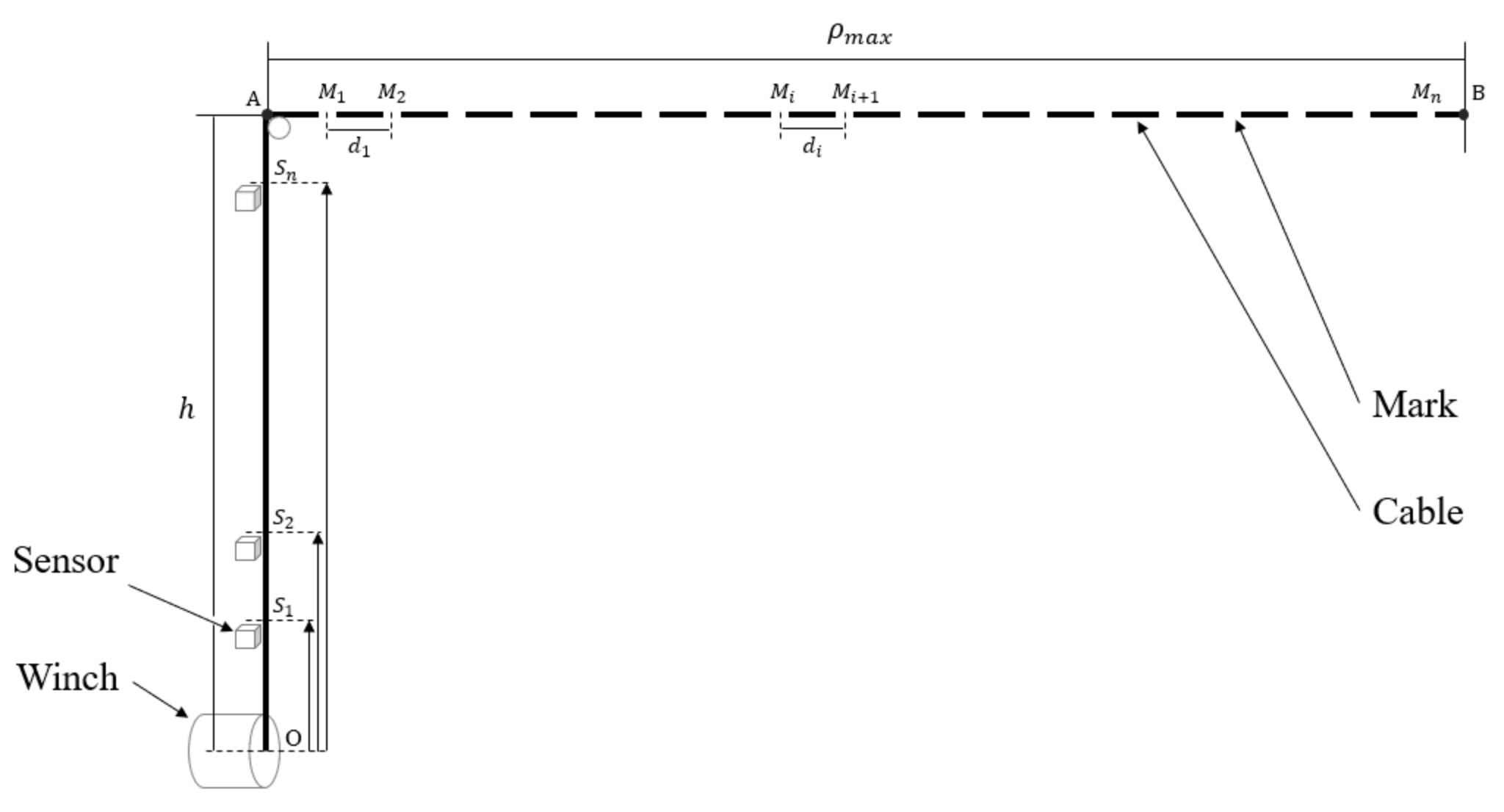}
  \caption{Instrumented cable}
  \label{fig:f2}
\end{figure}

For synthetic cable instrumentation, we will place metallic marks of small dimension\footnote{For example, width of 1 cm.} (Figure~\ref{fig:f2}).The length $\mathbf{\|BM_i\|}$ between mark $M_i$ and distal point B is known, as well as sensor positions in the reference frame. The distance $d_i$ between two successive marks will be defined from a set of known values.

For maintenance reasons, a motorized winch will be placed on the ground. To wind cable through support, we use guide pulleys; cable will be wound from the winch center O to mobile platform (point B), passing by top of support (point A). $\rho_{max}$ is maximum $\mathbf{\|AB\|}$ length, total cable length $l_max$ we will need:
\begin{equation}\label{eq:eq0}
  l_{max}=h+\rho_{max}
\end{equation}
When winding or unwinding instrumented cable, sensors will indicate the passage of a mark and time of this passage. If we can identify which mark is passing in front of the sensor, then we know cable length $\mathbf{\rho = \|AB\|}$.

\section{Notion of event and possible events}
  \subsection{Notion of event}

An event $E_{i,j}$ is detection of a mark $M_i$, by a sensor $S_j$, at an instant $t_{i,j}$. At detection time, the length $\mathbf{\|AB\|}$ will be noted $\rho_{i,j}$:
\begin{equation}\label{eq:eq1}
  E_{i,j} = \{t_{i,j}, M_i, S_j, \rho_{i,j} \}
\end{equation}

\begin{figure}[!h]
  \centering
  \includegraphics[width=.6\linewidth]{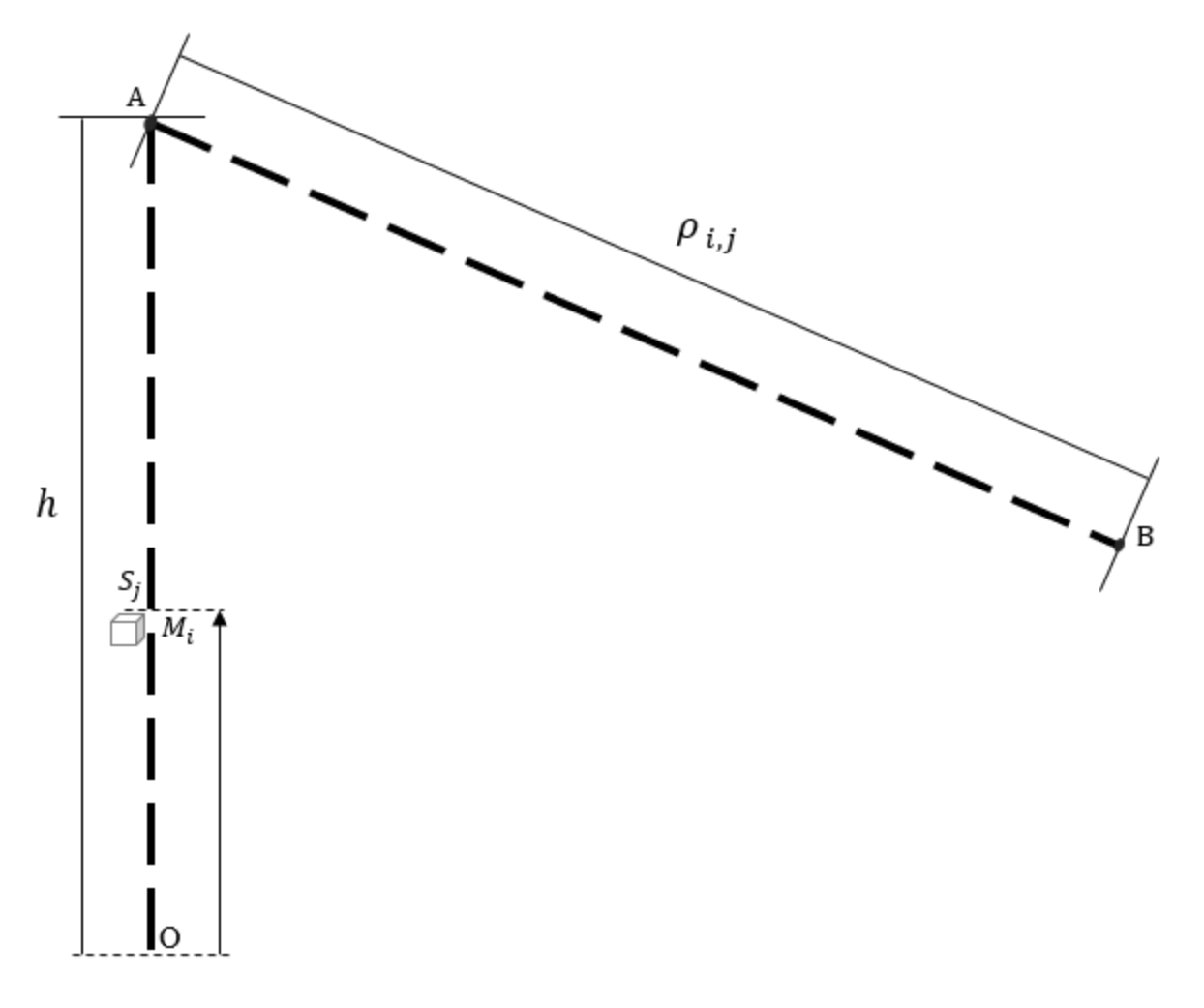}
  \caption{Configuration during passage of mark $M_i$ by sensor $S_j$}
  \label{fig:f3}
\end{figure}

Let $h$ be support height, $\mathbf{\|OS_j\|}$ sensor $S_j$ vertical position from winch center and $\mathbf{\|BM_i\|}$ distance between mark $M_i$ and distal point B (Figure~\ref{fig:f3}).
\begin{equation}\label{eq:eq2}
  \rho_{i,j} = \mathbf{\|BM_i\|} - (h - \mathbf{\|OS_j\|})
\end{equation}

The difference in cable length between two successive marks detection ($E_{i,j}$ and $E_{k,l}$), in increasing time:
\begin{equation}\label{eq:eq3}
  \Delta\rho = |\rho_{i,j}(t+\Delta t) - \rho_{k,l}(t)|
\end{equation}

When CDPR starts, we cannot determine when the first detection will take place, or on which sensor, or which mark is detected; but the following detections will provide information. We will present how to use this information, both to obtain $\rho_{i,j}$ and calibrate cables length.

\subsection{Identification of possible events}

To begin, we suppose that all CDPR parameters are known: maximum $\mathbf{\|AB\|}$ length $\rho_{max}$, number of marks $n_m$, distances $\mathbf{\|BM_i\|}$ between successive marks, height of supports $h$, number of stations $n_s$ and their vertical position from winch center $\mathbf{\|OS_j\|}$. The maximum number of possible events $n_e$ is given by:
\begin{equation}\label{eq:eq4}
  n_e = n_m \times n_s
\end{equation}

To identify \emph{possible events}, we will use following method:
\begin{enumerate}
\item at start, cable is at its maximum length $\rho_{max}$, it is wound up at a constant and known speed $v$;
\item we calculate time $t_{i,j}$, when mark $M_i$ is detected by sensor $S_j$, this for all possible pairs $(M_i, S_j)$;
\item we order events $E_{i,j}$ in increasing time.
\end{enumerate}

At a given time $t$, we have: 
\begin{equation}\label{eq:eq5}
  \mathbf{\|AB\|} = \rho_{max} - v.t
\end{equation}

In addition, we have: 
\begin{equation}\label{eq:eq6}
  \mathbf{\|OM_i\|} = \mathbf{\|OA\|} + \mathbf{\|AB\|} - \mathbf{\|BM_i\|}
\end{equation}

Which corresponds to:
\begin{equation}\label{eq:eq7}
  \mathbf{\|OM_i\|} = h + \rho_{max} - v.t - \mathbf{\|BM_i\|}
\end{equation}

That we can simplify, using equation (\ref{eq:eq0}):
\begin{equation}
  \mathbf{\|OM_i\|} = l_{max} - v.t - \mathbf{\|BM_i\|}
\end{equation}

When detecting mark $M_i$ by sensor $S_j$, we must have:
\begin{equation}\label{eq:eq8}
  \mathbf{\|OM_i\|} = \mathbf{\|OS_j\|}
\end{equation}

Hence detection time:
\begin{equation}\label{eq:eq9}
  t_{i,j} = \frac{l_{max} - \mathbf{\|BM_i\|} - \mathbf{\|OS_j\|}}{v}
\end{equation}
\begin{center}
with, $i \in [|1,n_m|]$ and $j \in [|1,n_s|]$
\end{center}
By determining this detection time for all pairs $(M_i, S_j)$, we will obtain the set of possible events $\{E_{1,n}, \ldots, E_{i,j}, \ldots, E_{n,1}\}$, both in terms of time and $\Delta\rho$ variation.

\begin{figure}[!h]
  \centering
  \includegraphics[width=1\linewidth]{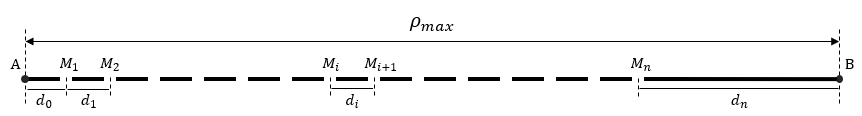}
  \caption{Placement of marks on cable}
  \label{fig:f41}
\end{figure}

When we are at maximum length $\rho_{max}$, the cable is completely unwound. For calibration, we can only wind the cable. Therefore, it is useless to put marks on segment cable $\overline\mathbf{OA}$ (Figure~\ref{fig:f41}).
\clearpage
\begin{center}
with, $d_0 = \mathbf{\|AM_1\|} = \mathbf{\|AS_n\|}$ and $d_n = \mathbf{\|BM_n\|}$
\end{center}

In order to avoid first mark $M_1$, or last sensor $S_n$, being superposed on point A, we define $d_0$, such that:
\begin{equation}\label{eq:eqc1}
  d_0 = \min_{[|1,n_m|]}(d_i)
\end{equation}
This first condition (\textbf{C.1}) allows to place mark $M_1$, and sensor $S_n$, at a minimum distance $d_0$ from A. This distance must be different from zero and two successive marks must not superpose. For this, we define the following second condition (\textbf{C.2}):
\begin{equation}\label{eq:eqcc2}
  \forall i \in [|1,n_m|], d_i \ne 0
\end{equation}

Calibration requires training a minimum of cable. The last event, by equation (\ref{eq:eq1}), will be $E_{n,1} = \{t_{n,1}, M_n, S_1, \rho_{n,1} \}$.
\begin{figure}[!h]
  \centering
  \includegraphics[width=.4\linewidth]{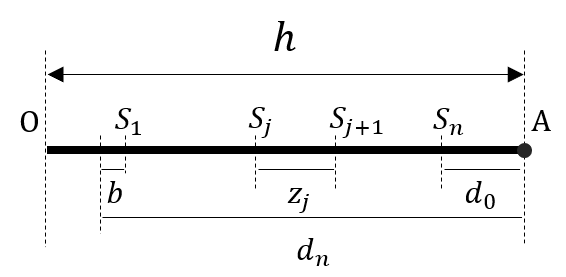}
  \caption{Placement of sensors on support}
  \label{fig:f42}
\end{figure}
To be able to wind cable until last event, without two points A and B superposition, we define a bost $b$ (Figure~\ref{fig:f42}) such as (\textbf{C.3}):
\begin{equation}\label{eq:eqc2}
   h - \mathbf{\|OS_1\|} - d_n + b = 0
\end{equation}
It is necessary to fulfill previous condition (\textbf{C.3}), to prevent mobile platform from being on support. Similarly, to avoid the superposition of two successive sensors, we define the following fourth condition (\textbf{C.4}):
\begin{equation}\label{eq:eqcc4}
  \forall j \in [|1,n_s|], z_j \ne 0
\end{equation}

It is therefore useless  to put marks on $\overline\mathbf{AM_1}$ and $\overline\mathbf{BM_n}$ cable segments. In the next \emph{examples}, we will instrument only cable segment $\overline\mathbf{M_1M_n} = \overline\mathbf{OB} \cap (\overline\mathbf{OA} \cup \overline\mathbf{AM_1} \cup \overline\mathbf{BM_n})$. The purpose of examples is to learn how to realise this type of \emph{autocalibration subsystem}.

\subsection{Examples}
We will analyze several examples, to define more formulas and conditions for the implementation of this autocalibration subsystem. For all examples, we fixe winding speed at $v = 1 m.s^{-1}$. Speed is the division of displacement by time. For the same distance, variation in speed will influence only detection time.

\subsubsection{Medium CDPR with two sensors}
\begin{figure}[!h]
  \centering
  \includegraphics[width=.3\linewidth]{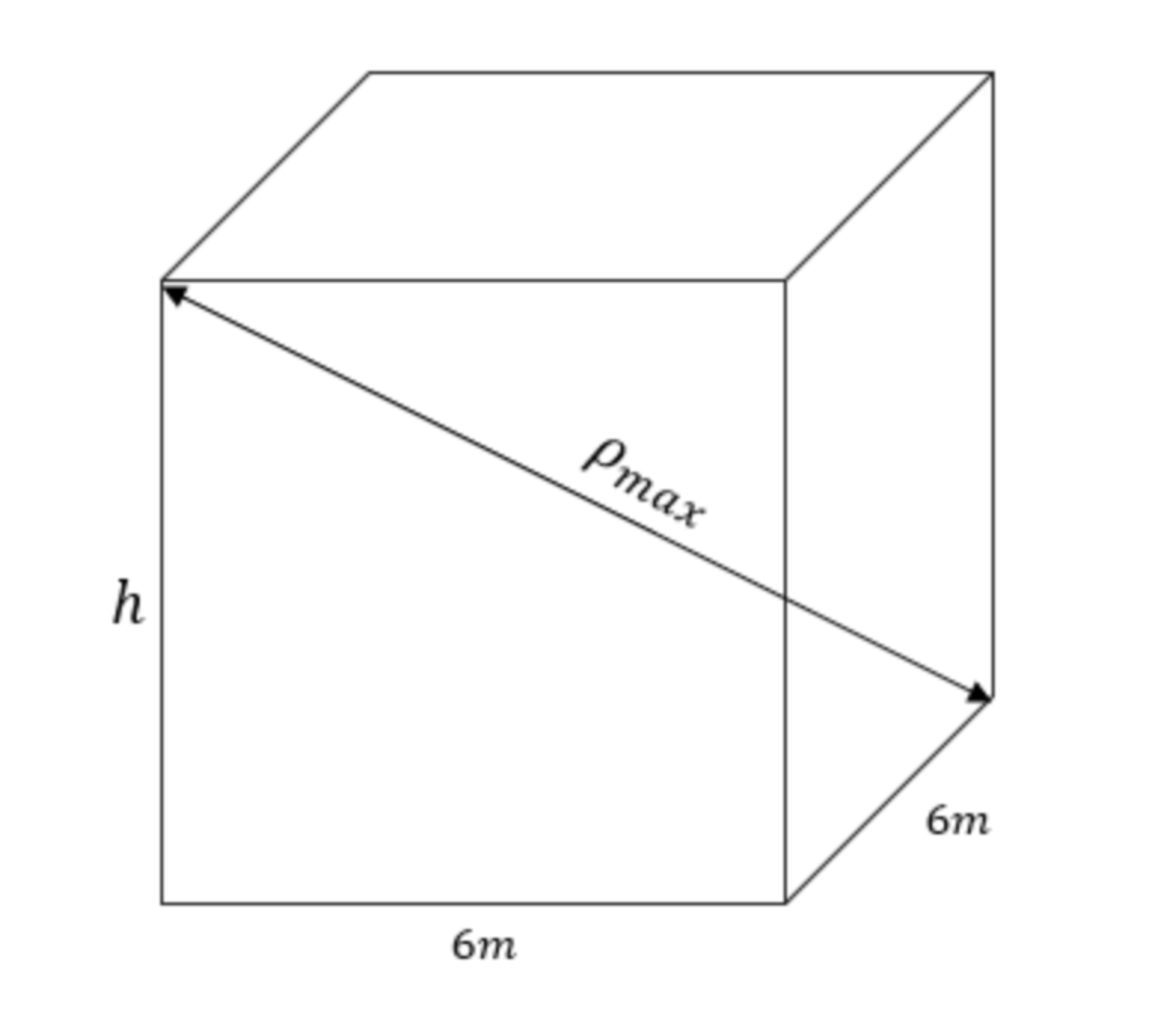}
  \caption{Medium CDPR congestion}
  \label{fig:f1e11}
\end{figure}

A medium-sized CDPR, typically $h = 6 m$, covers a cube of $6\times6\times6 m$, with $\rho_{max} = 11m$ ($\sqrt{6^2 + 6^2 + 6^2} \approx 10.4$) (Figure~\ref{fig:f1e11}).
%\clearpage

\textbf{Remark 1:} for maximum cable length dimensioning, we are interested in the \emph{maximal workspace} $(height \times width \times length)$. In general, \emph{reachable workspace} is smaller than maximal one, due $-$for example$-$ to Manipulability~\cite{yoshikawa1985manipulability}, ~\cite{merlet2006jacobian}. Dimensioning cable length over maximum workspace will allow $-$for safety reasons$-$ to keep a minimum of cable in the drum.

The distance between successive marks will be constant: $\forall i \in [|1,n_m|], d_i = 1 m$. Then $d_0 = 1 m$ ($\min_{[|1,n_m|]}(d_i) = 1$), condition (C.1) is satisfied; first mark and $n^{th}$ sensor will be placed at a distance of $1 m$ from point A. To be able to use marks as close as possible to point B, which will allow us to be able to calibrate most of cable length, we suggest to place first sensor at \emph{third} of high $h$:
\begin{equation}\label{eq:eqos1}
  \mathbf{\|OS_1\|} = \frac{h}{3}
\end{equation}
We will therefore place sensors on remaining two thirds of support. First sensor will be placed on support, by previous equation, at:
\begin{center}
\fbox{$\mathbf{\|OS_1\|}= 2m$}
\end{center}
We boost cable length by $b = 1m$, with equation (\ref{eq:eqc2}):
\begin{equation}\label{eq:eqdn}
  d_n = h - \mathbf{\|OS_1\|} + b
\end{equation}
To satisfy (C.3) condition, we get:
\begin{center}
\fbox{$d_n= 5m$}
\end{center}
The distance between successive sensors is given by:
\begin{equation}
\forall j \in [|1,n_s|], z_j = \big|\mathbf{\|OS_{j+1}\|}- \mathbf{\|OS_{j}\|}\big|
\end{equation}
In this example, distance $z_j$ will be constant:
\begin{center}
\fbox{$\forall j \in [|1,n_s|], z_j = 3 m$}
\end{center}
The number of sensors we can place is given by:
\begin{equation}\label{eq:eqns}
  n_s = 1 + \frac{h - d_{0} - \mathbf{\|OS_{1}\|}}{z_j}
\end{equation}
We start with one sensor, placed in a distance $\mathbf{\|OS_1\|}$ from O and in each support segment bound, we will place another sensor. The number of support segments $\overline\mathbf{S_{j}S_{j+1}}$ is given by $\frac{h - d_{0} - \mathbf{\|OS_{1}\|}}{z_j}$. We round value found to an integer\footnote{We cannot use half of sensor.}. Then $n_s = 2$, they will be placed, vertically on support, as follow:
\begin{table}[!h]
  \begin{center}
    \caption{$1^{st}$ example sensors position}
    \medskip
      \label{tab:t1}
    \begin{tabular}{|l|r|}
      \hline
      Sensor &  $\mathbf{\|OS_j\|} \hspace{.1in} [m]$\\
      \hline
      $S_2$  &  5\\
      \hline
      $S_1$  &  2\\
      \hline
    \end{tabular}
  \end{center}
\end{table}

The number of marks we can use is given by:
\begin{equation}\label{eq:eqnm}
  n_m = 1 + \frac{\rho_{max} - d_{0} - d_{n}}{d_i}
\end{equation}
Similarly, we start with one mark, placed in a distance $d_0$ from A and in each cable segment bound, we will place another mark. The number of cable segments $\overline\mathbf{M_{i}M_{i+1}}$ is given by $\frac{\rho_{max} - d_{0} - d_{n}}{d_i}$. We round value found to an integer\footnote{We cannot use half of mark.}. Then $n_m = 6$, they will be placed on cable as follow:
\begin{table}[!h]
  \begin{center}
    \caption{$1^{st}$ example distance between successive marks}
    \medskip
      \label{tab:t11}
    \begin{tabular}{|l|r|}
      \hline
      Mark &  $\mathbf{\|BM_i\|} \hspace{.1in} [m]$\\
      \hline
      $M_1$  &  10\\
      \hline
      $M_2$  &  9\\
      \hline
      $M_3$  &  8\\
      \hline
      $M_4$  &  7\\
      \hline
      $M_5$  &  6\\
    \hline
      $M_6$  &  5\\
    \hline
    \end{tabular}
  \end{center}
\end{table}

\newpage

The number of possible events, by equation (\ref{eq:eq4}), $n_e = 12$, bellow is list of all possible events:
\begin{table}[!h]
  \begin{center}
    \caption{$1^{st}$ example list of possible events}
    \medskip
      \label{tab:t12}
    \begin{tabular}{|c|c|c|c|c|}
      \hline
      $t_{i,j} \hspace{.1in} [s]$ & $M_i$ & $S_j$ & $\rho_{i,j} \hspace{.1in} [m]$ & $\Delta\rho \hspace{.1in} [m]$\\
      \hline
      2.00 & $M_1$ & $S_2$ & 9.00 & \\
      \hline
      3.00 & $M_2$ & $S_2$ & 8.00 & 1.00\\
      \hline
      4.00 & $M_3$ & $S_2$ & 7.00 & 1.00\\
      \hline
      5.00 & $M_1$ & $S_1$ & 6.00 & 1.00\\
      \hline
      5.00 & $M_4$ & $S_2$ & 6.00 & 0.00\\
      \hline
      6.00 & $M_2$ & $S_1$ & 5.00 & 1.00\\
      \hline
      6.00 & $M_5$ & $S_2$ & 5.00 & 0.00\\
      \hline
      7.00 & $M_3$ & $S_1$ & 4.00 & 1.00\\
      \hline
      7.00 & $M_6$ & $S_2$ & 4.00 & 0.00\\
      \hline
      8.00 & $M_4$ & $S_1$ & 3.00 & 1.00\\
      \hline
      9.00 & $M_5$ & $S_1$ & 2.00 & 1.00\\
      \hline
      10.00 & $M_6$ & $S_1$ & 1.00 & 1.00\\
      \hline
    \end{tabular}
  \end{center}
\end{table}

Certain events are simultaneous. This simultaneity is due to detection of two different marks by two different sensors, at same time. For $t_{i,j}=t_{k,l}$, using equation (\ref{eq:eq9}), we get:
\begin{equation}\label{eq:eq10}
  \frac{l_{max} - \mathbf{\|BM_i\|} - \mathbf{\|OS_j\|}}{v} = \frac{l_{max} - \mathbf{\|BM_k\|} - \mathbf{\|OS_l\|}}{v}
\end{equation}
with $v \ne 0$, previous equation can be simplified as:
\begin{equation}\label{eq:eq11}
  \mathbf{\|BM_i\|} + \mathbf{\|OS_j\|} = \mathbf{\|BM_k\|} + \mathbf{\|OS_l\|}
\end{equation}
that we can write, using equation (\ref{eq:eq2}), as:
\begin{equation}\label{eq:eq13}
  \mathbf{\|BM_i\|} - (h - \mathbf{\|OS_j\|}) = \mathbf{\|BM_k\|} - (h - \mathbf{\|OS_l\|})
\end{equation}
which corresponds to:
\begin{equation}\label{eq:eq14}
  |\rho_{i,j}(t_{i,j}) - \rho_{k,l}(t_{k,l})| = 0
\end{equation}
However, each instant $t$, must be linked to a \emph{unique event}. Which leads us to following condition (\textbf{C.5}), by equation (\ref{eq:eq3}):

$\forall i, k \in [|1,n_m|]$ and $\forall j, l \in [|1,n_s|]$,
\begin{equation}\label{eq:eq12}
  \Delta\rho \ne 0
\end{equation}
To sort, each time, we will keep pair $(M_i,S_j)$ with \emph{smallest} index $i$ and \emph{biggest} index $j$, otherwise $\min(\frac{i}{j})$; in case of equality, we favor pair with $\min(i)$. This amounts to taking shortest distance for calibration.\\

%\newpage

After taking into account condition (C.5), precedent table becomes:
\begin{table}[!h]
  \begin{center}
    \caption{$1^{st}$ example list of possible events (after rectification)}
    \medskip
      \label{tab:t13}
    \begin{tabular}{|c|c|c|c|c|}
      \hline
      $t_{i,j} \hspace{.1in} [s]$ & $M_i$ & $S_j$ & $\rho_{i,j} \hspace{.1in} [m]$ & $\Delta\rho \hspace{.1in} [m]$\\
      \hline
      2.00 & $M_1$ & $S_2$ & 9.00 & \\
      \hline
      3.00 & $M_2$ & $S_2$ & 8.00 & 1.00\\
      \hline
      4.00 & $M_3$ & $S_2$ & 7.00 & 1.00\\
      \hline
      5.00 & $M_1$ & $S_1$ & 6.00 & 1.00\\
      \hline
      6.00 & $M_2$ & $S_1$ & 5.00 & 1.00\\
      \hline
      7.00 & $M_3$ & $S_1$ & 4.00 & 1.00\\
      \hline
      8.00 & $M_4$ & $S_1$ & 3.00 & 1.00\\
      \hline
      9.00 & $M_5$ & $S_1$ & 2.00 & 1.00\\
      \hline
      10.00 & $M_6$ & $S_1$ & 1.00 & 1.00\\
      \hline
    \end{tabular}
  \end{center}
\end{table}

\begin{figure}[!h]
  \centering
  \includegraphics[width=.6\linewidth]{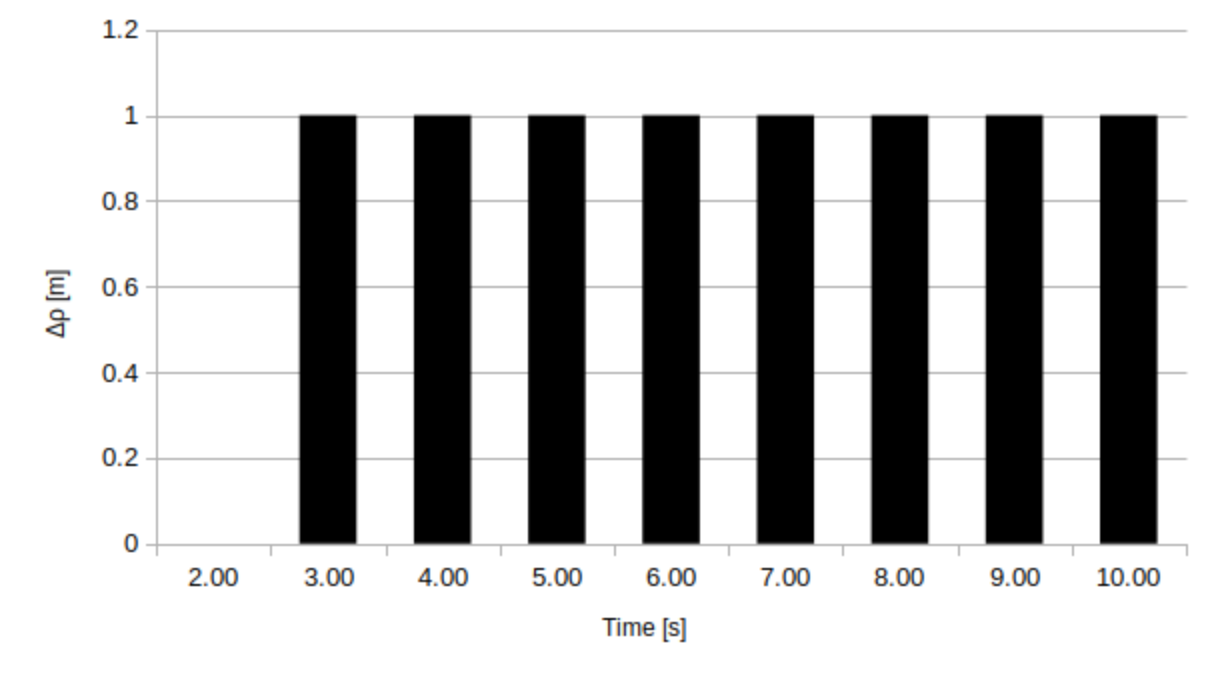}
  \caption{$1^{st}$ example $\Delta\rho(t)$}
  \label{fig:f2e11}
\end{figure}

The fifth condition (C.5) reduces number of exploitable events to $n_e=9$. With a winding speed $v = 1m.s^{-1}$, we needed $t = 2.00s$, to have first event $E_{1,2}$, which corresponds to a distance of $2m$ $(1m.s^{-1}\times 2s)$; the distance found is consistent with value of $2*d_0$. Last event $E_{6,1}$ at $t_{6,1}=10.00s$ and $\rho_{6,1}=1m$, which indicates that mobile platform $-$ by the end $-$ is about $1 m$ from support. Value of $\Delta\rho$ is constant as a function of time (Figure~\ref{fig:f2e11}).

\clearpage

Mean value $\overline{\Delta\rho}$ of $\Delta\rho(t)$ is defined by:
\begin{equation}\label{eq:eqmeanro}
\overline{\Delta\rho} = \frac{\sum_{e=2}^{n_e}\Delta\rho}{n_e-1}
\end{equation}

Index $e$ designates \emph{event}. $n_e$ is number of events with respect to fifth condition (C.5). We start computing $\Delta\rho$ from second event, until last one. $\overline{\Delta\rho}$ value  gives us an idea about cable length we must wound $-$on average$-$ for a new detection. In this example, $\overline{\Delta\rho}=1m$.\\

Standard deviation $\sqrt{var(\Delta\rho)}$ of $\Delta\rho(t)$ is defined by:
\begin{equation}\label{eq:eqsqrtvar}
\sqrt{var(\Delta\rho)} = \sqrt{\frac{\sum_{e=2}^{n_e}(\Delta\rho-\overline{\Delta\rho})^2}{n_e-2}}
\end{equation}
$\sqrt{var(\Delta\rho)}$ gives us an idea of difference in cable length that we will have $-$on average$-$ between two successive detections. In this example, $\sqrt{var(\Delta\rho)}=0m$, it means there is no difference; which in coherence with $\Delta\rho=f(t)$ function values.\\

Thereafter, we will vary $d_i$ distance, then analyze again $\Delta\rho(t)$. For all the next examples, the five conditions (C.1), (C.2), (C.3), (C.4), (C.5) will be checked.

\subsubsection{Large CDPR with three sensors}
A large-sized CDPR, typically $h = 12 m$, covers a cube of $12\times12\times12 m$, \\ with $\rho_{max} = 21 m$ ($\sqrt{12^2 + 12^2 + 12^2} \approx 20.8$).

In this example, the distance between successive marks will be variable: \\ $\forall i \in [|1,n_m|], d_i \in \{0.5, 0.75, 1.00, 1.25, 1.50\} [m]$. The mean value $\bar d$ of $d_i$ is given by: 
\begin{equation}\label{eq:eqmeand}
  \bar{d}= \frac{\sum_{i=1}^{n_m}d_i}{n_m}
\end{equation}
We define the list of $d_i$ such that:
\begin{center}
\fbox{$\bar{d}= 1m$}
\end{center}

Then $d_0 = 0.5 m$, first mark and $n^{th}$ sensor will be placed at a distance of $0.5 m$ from point A. First sensor will be placed on support, by equation (\ref{eq:eqos1}), at:
\begin{center}
\fbox{$\mathbf{\|OS_1\|}= 4m$}
\end{center}
We boost cable length by $b = 1m$, by equation (\ref{eq:eqdn}), we get:
\begin{center}
\fbox{$d_n= 9m$}
\end{center}
The distance between successive sensors will be constant:
\begin{center}
$\forall j \in [|1,n_s|], z_j = 3.75 m$
\end{center}
The number of sensors we can place, by equation (\ref{eq:eqns}):
\begin{center}
\fbox{$n_s= 3$}
\end{center}
They will be placed, vertically on the support, as follow:

\begin{table}[!h]
  \begin{center}
    \caption{$2^{nd}$ example sensors position}
	\medskip
      \label{tab:t1}
    \begin{tabular}{|l|r|}
      \hline
      Sensor &  $\mathbf{\|OS_j\|} \hspace{.1in} [m]$\\
      \hline
      $S_3$  &  11.50\\
      \hline
      $S_2$  &  7.75\\
      \hline
      $S_1$  &  4.00\\
      \hline
    \end{tabular}
  \end{center}
\end{table}

For the number of marks, equation (\ref{eq:eqnm}) becomes:
\begin{equation}\label{eq:eqnm2}
  n_m = 1 + \frac{\rho_{max} - d_{0} - d_{n}}{\bar d}
\end{equation}

Then $n_m = 13$, they will be placed on cable as follow:
\begin{table}[!h]
  \begin{center}
    \caption{$2^{nd}$ example distance between successive marks}
	\medskip
      \label{tab:t21}
    \begin{tabular}{|l|r|}
      \hline
      Mark &  $\mathbf{\|BM_i\|} \hspace{.1in} [m]$\\
      \hline
      $M_1$  &  20.50\\
      \hline
      $M_2$  &  19.75\\
      \hline
      $M_3$  &  18.75\\
      \hline
      $M_4$  &  17.50\\
      \hline
      $M_5$  &  16.00\\
      \hline
      $M_6$  &  15.50\\
      \hline
      $M_7$  &  14.75\\
      \hline
      $M_8$  &  13.75\\
      \hline
      $M_9$  &  12.50\\
      \hline
      $M_{10}$  &  11.00\\
      \hline
      $M_{11}$  &  10.25\\
      \hline
      $M_{12}$  &  9.75\\
      \hline
      $M_{13}$  &  9.00\\
      \hline
    \end{tabular}
  \end{center}
\end{table}

%\clearpage

Distance $d_i$ is defined, for each mark, from $\{0.5, 0.75, 1.00, 1.25, 1.50\}$, in same order. We start with $d_0 = 0.5m$, after $1.50m$, next mark is placed at $d_i = 0.5$; except for last three marks, where we choose $d_i$ in such way to respect condition (C.3) $(d_n =\mathbf{\|BM_n\|})$. Reorganization of $d_i$ causes no significant changes.

Below is the list of possible events:
\begin{table}[!h]
  \begin{center}
    \caption{$2^{nd}$ example list of possible events}
	\medskip
      \label{tab:t22}
    \begin{tabular}{|c|c|c|c|c|}
      \hline
      $t_{i,j} \hspace{.1in} [s]$ & $M_i$ & $S_j$ & $\rho_{i,j} \hspace{.1in} [m]$ & $\Delta\rho \hspace{.1in} [m]$\\
      \hline
      1.00 & $M_1$ & $S_3$ & 20.00 & \\
      \hline
      1.75 & $M_2$ & $S_3$ & 19.25 & 0.75\\
      \hline
      2.75 & $M_3$ & $S_3$ & 18.25 & 1.00\\
      \hline
      4.00 & $M_4$ & $S_3$ & 17.00 & 1.25\\
      \hline
      4.75 & $M_1$ & $S_2$ & 16.25 & 0.75\\
      \hline
      $\cdots$ & $\cdots$ & $\cdots$ & $\cdots$ & $\cdots$\\
      \hline
      16.50 & $M_9$ & $S_1$ & 4.50 & 0.25\\
      \hline
      18.00 & $M_{10}$ & $S_1$ & 3.00 & 1.50\\
      \hline
      18.75 & $M_{11}$ & $S_1$ & 2.25 & 0.75\\
      \hline
      19.25 & $M_{12}$ & $S_1$ & 1.75 & 0.50\\
      \hline
      20.00 & $M_{13}$ & $S_1$ & 1.00 & 0.75\\
      \hline
    \end{tabular}
  \end{center}
\end{table}

\begin{figure}[!h]
  \centering
  \includegraphics[width=.7\linewidth]{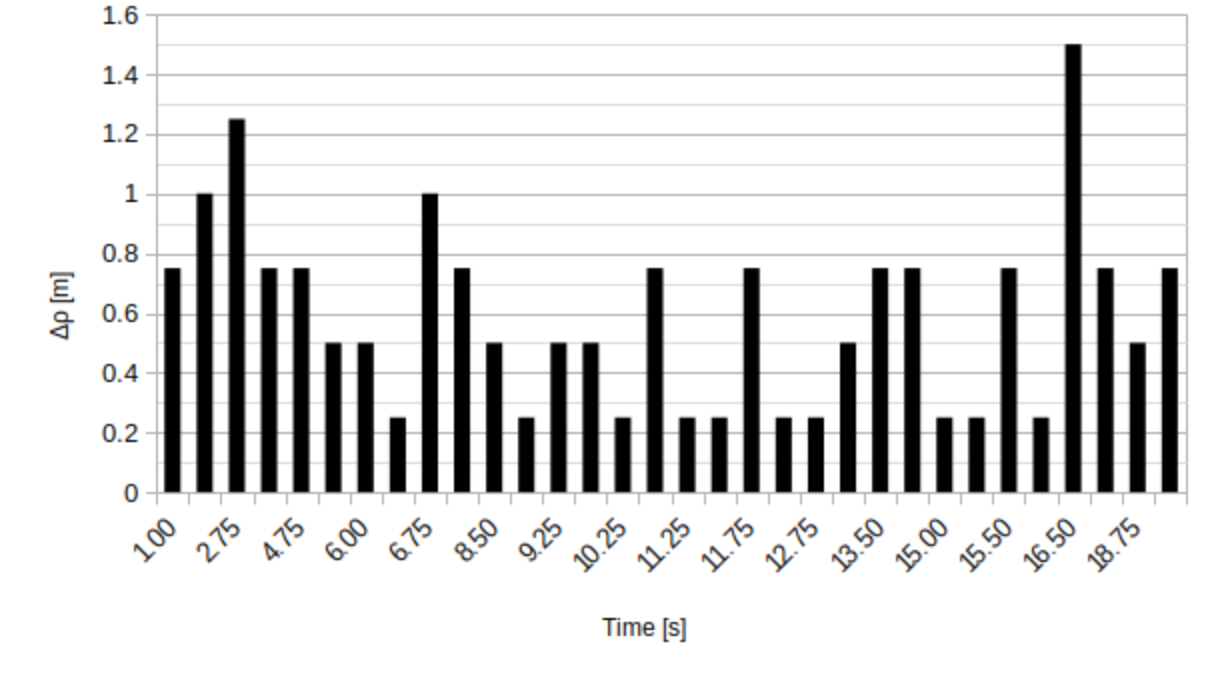}
  \caption{$2^{nd}$ example $\Delta\rho(t)$}
  \label{fig:f1e21}
\end{figure}

The total number of events, with respect to the three conditions $n_e=33$. First event $E_{1,3}$ at $t_{1,3} = 1.00s$ with $\rho_{1,3}=20.00m$. Last event $E_{13,1}$ at $t_{13,1}=20.00s$ with $\rho_{13,1}=1m$, mobile platform is about $1 m$ from support. Value of $\Delta\rho$ is variable as function of time (Figure~\ref{fig:f1e21}).

Variation of $d_i$ values leads to a variation of $\Delta\rho$. Mean value $\overline{\Delta\rho}=0.5m$, this value is reduced by $50\%$ compared to previous example, using a list of variable $d_i$ with $\bar d=1m$. Standard variation $\sqrt{var(\Delta\rho)}=0.3m$, which means that difference between successive $\Delta\rho$ values is around $0.3m$; contrary to previous example, this new configuration allows distinction between events. The distance between successive marks must therefore be variable. For this reason, we define the following sixth condition (\textbf{C.6}):
\begin{equation}\label{eq:eqcc6}
  \forall i \in [|1,n_m|], d_{i+1} - d_i \ne 0
\end{equation}
In the next example, we will vary not only the distance between successive marks but also the distance between successive sensors.

\subsubsection{Very large CDPR with five sensors}
A large-sized CDPR, typically $h = 18 m$, covers a cube of $18\times18\times18 m$, \\ with $\rho_{max} = 32 m$ ($\sqrt{18^2 + 18^2 + 18^2} \approx 31.2$).\\
The distance between successive marks will be variable: \\ $\forall i \in [|1,n_m|], d_i \in \{0.25, 0.50, 0.75, 1.00, 1.25, 1.50, 1.75, 2.00, 2.25, 2.5\} [m]$. The mean value $\bar d = 1.37m$. Then $d_0 = 0.25 m$, first mark and $n^{th}$ sensor will be placed at a distance of $0.25 m$ from point A. First sensor will be placed on support, by equation (\ref{eq:eqos1}), at:
\begin{center}
\fbox{$\mathbf{\|OS_1\|}= 6m$}
\end{center}
We boost cable length by $b = 1m$, by equation (\ref{eq:eqdn}), we get:
\begin{center}
\fbox{$d_n= 13m$}
\end{center}

\textbf{Remark 2:} at last event, cable length corresponds to value $b$, used to \emph{boost} cable. This will avoid having the mobile platform on support.\\

The distance between successive sensors will be variable:\\
$\forall j \in [|1,n_s|], z_j \in \{1, 2, 3, 5\} [m]$ \\
The mean value $\bar z$ of $z_j$ is given by: 
\begin{equation}\label{eq:eqmeanz}
  \bar{z}= \frac{\sum_{j=1}^{n_s}z_j}{n_s}
\end{equation}
We defined the list of $z_j$ such that:
\begin{center}
\fbox{$\bar{z}= 2.75m$}
\end{center}
For the number of sensors we can place, equation (\ref{eq:eqns}) becomes:
\begin{equation}\label{eq:eqzn}
  n_s = 1 + \frac{h - d_{0} - \mathbf{\|OS_{1}\|}}{\bar z}
\end{equation}
Then $n_s=5$, they will be placed, vertically on the support, as follow:
\begin{table}[!h]
  \begin{center}
    \caption{$3^{rd}$ example sensors position}
	\medskip
      \label{tab:t1}
    \begin{tabular}{|l|r|}
      \hline
      Sensor &  $\mathbf{\|OS_j\|} \hspace{.1in} [m]$\\
      \hline
      $S_5$  &  17.75\\
      \hline
      $S_4$  &  16.00\\
      \hline
      $S_3$  &  14.00\\
      \hline
      $S_2$  &  11.00\\
      \hline
      $S_1$  &  6.00\\
      \hline
    \end{tabular}
  \end{center}
\end{table}

The number of marks, by equation (\ref{eq:eqnm2}):
\begin{center}
\fbox{$n_m= 14$}
\end{center}

\clearpage
They will be placed on cable as follow:
\begin{table}[!h]
  \begin{center}
    \caption{$3^{rd}$ example distance between successive marks}
	\medskip
      \label{tab:t21}
    \begin{tabular}{|l|r|}
      \hline
      Mark &  $\mathbf{\|BM_i\|} \hspace{.1in} [m]$\\
      \hline
      $M_1$  &  31.75\\
      \hline
      $M_2$  &  31.25\\
      \hline
      $M_3$  &  30.50\\
      \hline
      $M_4$  &  29.50\\
      \hline
      $M_5$  &  28.25\\
      \hline
      $M_6$  &  26.75\\
      \hline
      $M_7$  &  25.00\\
      \hline
      $M_8$  &  23.00\\
      \hline
      $M_9$  &  20.75\\
      \hline
      $M_{10}$  &  18.25\\
      \hline
      $M_{11}$  &  18.00\\
      \hline
      $M_{12}$  &  16.75\\
      \hline
      $M_{13}$  &  15.25\\
      \hline
      $M_{14}$  &  13.00\\
      \hline
    \end{tabular}
  \end{center}
\end{table}

Using this configuration, the list of possible events is:
\clearpage
\begin{table}[!h]
  \begin{center}
    \caption{$3^{rd}$ example list of possible events}
	\medskip
      \label{tab:t22}
    \begin{tabular}{|c|c|c|c|c|}
      \hline
      $t_{i,j} \hspace{.1in} [s]$ & $M_i$ & $S_j$ & $\rho_{i,j} \hspace{.1in} [m]$ & $\Delta\rho \hspace{.1in} [m]$\\
      \hline
      0.50 & $M_1$ & $S_5$ & 31.50 & \\
      \hline
      1.00 & $M_2$ & $S_5$ & 31.00 & 0.5\\
      \hline
      1.75 & $M_3$ & $S_5$ & 30.25 & 0.75\\
      \hline
      2.25 & $M_1$ & $S_4$ & 29.75 & 0.5\\
      \hline
      2.75 & $M_2$ & $S_4$ & 29.25 & 0.5\\
      \hline
      $\cdots$ & $\cdots$ & $\cdots$ & $\cdots$ & $\cdots$\\
      \hline
      25.75 & $M_10$ & $S_1$ & 6.25 & 2\\
      \hline
      26.00 & $M_{14}$ & $S_2$ & 6.00 & 0.25\\
      \hline
      27.25 & $M_{12}$ & $S_1$ & 4.75 & 1.25\\
      \hline
      28.75 & $M_{13}$ & $S_1$ & 3.25 & 1.50\\
      \hline
      31.00 & $M_{14}$ & $S_1$ & 1.00 & 2.25\\
      \hline
    \end{tabular}
  \end{center}
\end{table}

\begin{figure}[!h]
  \centering
  \includegraphics[width=.7\linewidth]{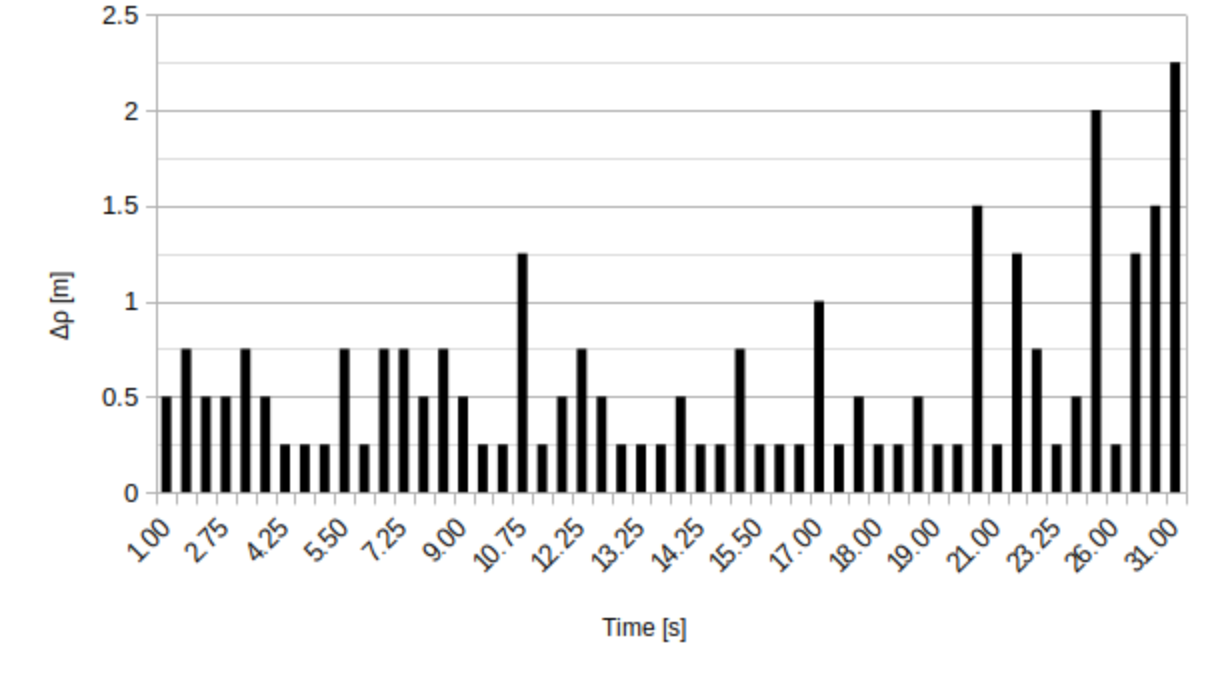}
  \caption{$3^{rd}$ example $\Delta\rho(t)$}
  \label{fig:f1e31}
\end{figure}

The total number of events, with respect to the three conditions $n_e=53$. First event $E_{1,5}$ at $t_{1,5} = 0.50s$ with $\rho_{1,5}=31.50m$. Last event $E_{14,1}$ at $t_{13,1}=31.00s$ with $\rho_{14,1}=1m$, mobile platform is about $1 m$ from support. The function $\Delta\rho=f(t)$ is variable (Figure~\ref{fig:f1e31}).

Cable length average value that must be wound, for a new detection, is $\overline{\Delta\rho}=0.5m$. Variation of $d_i$ and $z_j$ leads to increase variation between $\Delta\rho$ successive values, standard variation $\sqrt{var(\Delta\rho)}=0.5m$, with this new configuration, we will have \emph{one new and distinct detection, every half meter}. The distance between successive sensors must therefore be variable. Which brings us to the seventh condition (\textbf{C.7}):
\begin{equation}\label{eq:eqcc7}
  \forall j \in [|1,n_s|], z_{j+1} - z_j \ne 0
\end{equation}
The maximum value of $\Delta\rho$ is $2.5m$, which means that in some cases, we have to wind $2.5m$ for a new detection. Now, we will try to bring sensors as close as possible to winch center O. Sensors will, this time, be placed vertically on support as follows:

%\clearpage

\begin{table}[!h]
  \begin{center}
    \caption{$3^{rd}$ example sensors position (after changement)} 
	\medskip
      \label{tab:t1}
    \begin{tabular}{|l|r|}
      \hline
      Sensor &  $\mathbf{\|OS_j\|} \hspace{.1in} [m]$\\
      \hline
      $S_5$  &  17.75\\
      \hline
      $S_4$  &  11.00\\
      \hline
      $S_3$  &  9.00\\
      \hline
      $S_2$  &  7.00\\
      \hline
      $S_1$  &  6.00\\
      \hline
    \end{tabular}
  \end{center}
\end{table}

With this changement in sensor positioning, $\Delta\rho=f(t)$ function becomes (Figure~\ref{fig:f1e32}):

\begin{figure}[!h]
  \centering
  \includegraphics[width=.7\linewidth]{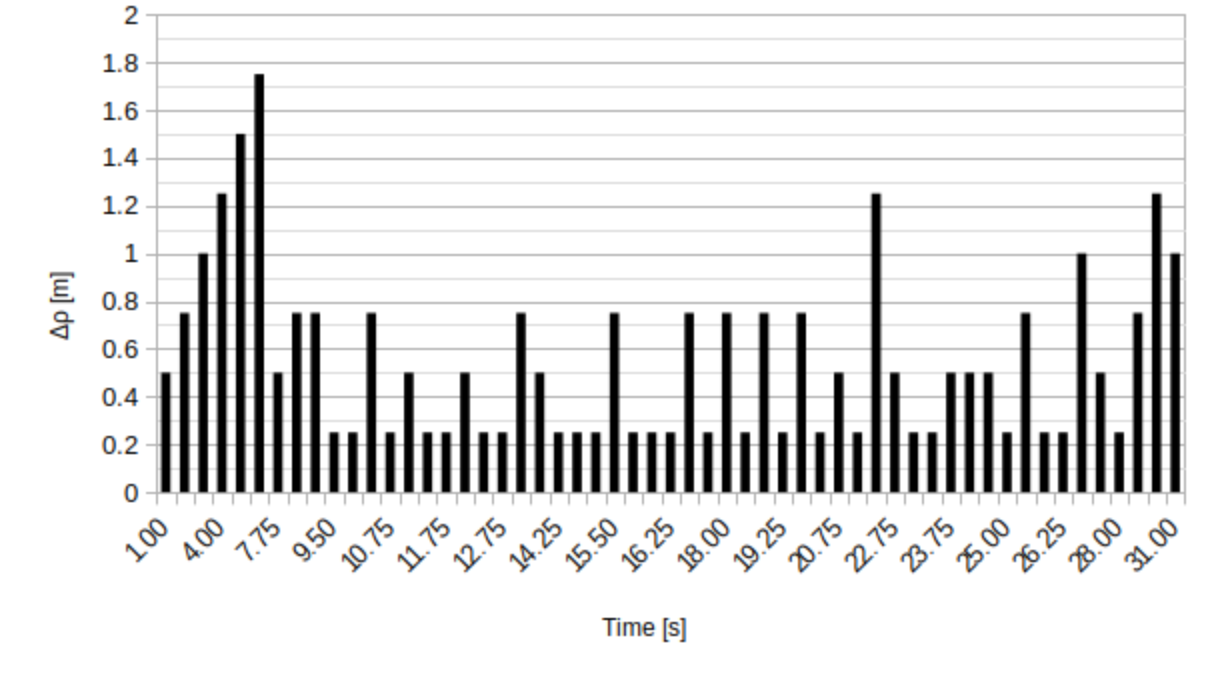}
  \caption{$3^{rd}$ example $\Delta\rho(t)$ (after changement)}
  \label{fig:f1e32}
\end{figure}

Bringing sensors closer to the winch center avoids a difference in cable length $\Delta\rho$ greater than or equal to $2m$. This reduces in cable length needed to wind $-$in some cases$-$ for a new detection, compared to the previous example. It would then be ingenious to place sensors as close as possible to winch center. And if we add even more sensors, we further reduce the difference in cable length between successive detections and increase the number of events. Autocalibration consists of winding cable until the acquisition of several distinct events.

\section{Calibration stroke}

\begin{figure}[!h]
  \centering
  \includegraphics[width=1\linewidth]{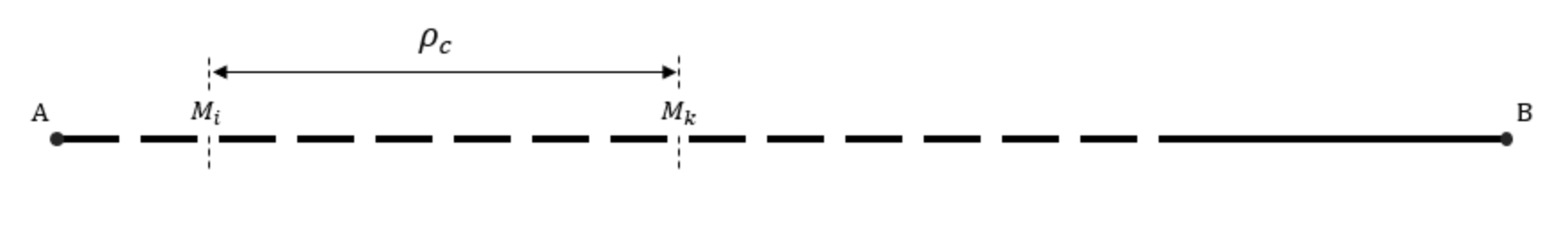}
  \caption{Calibration stroke}
  \label{fig:f7}
\end{figure}

The \emph{Calibration stroke} $\rho_c$ is cable length we must wound. Calibration consists of several events exploitation. The first event allows us to define length "zero". The $\rho_c$ length corresponds to the difference between the first and last events (resp. $E_{i,j}$ and $E_{k,l}$) (Figure~\ref{fig:f7}).

\begin{equation}\label{eq:eqrhoc}
  \rho_c = |\rho_{k,l}(t_{k,l}) - \rho_{i,j}(t_{i,j})|
\end{equation}
To minimize calibration stroke, we must reduce difference between last and first events. It is identification of these successive events that will allow auto-calibration.

\subsection{Autocalibration approach}
\begin{figure}[!h]
  \centering
  \includegraphics[width=.3\linewidth]{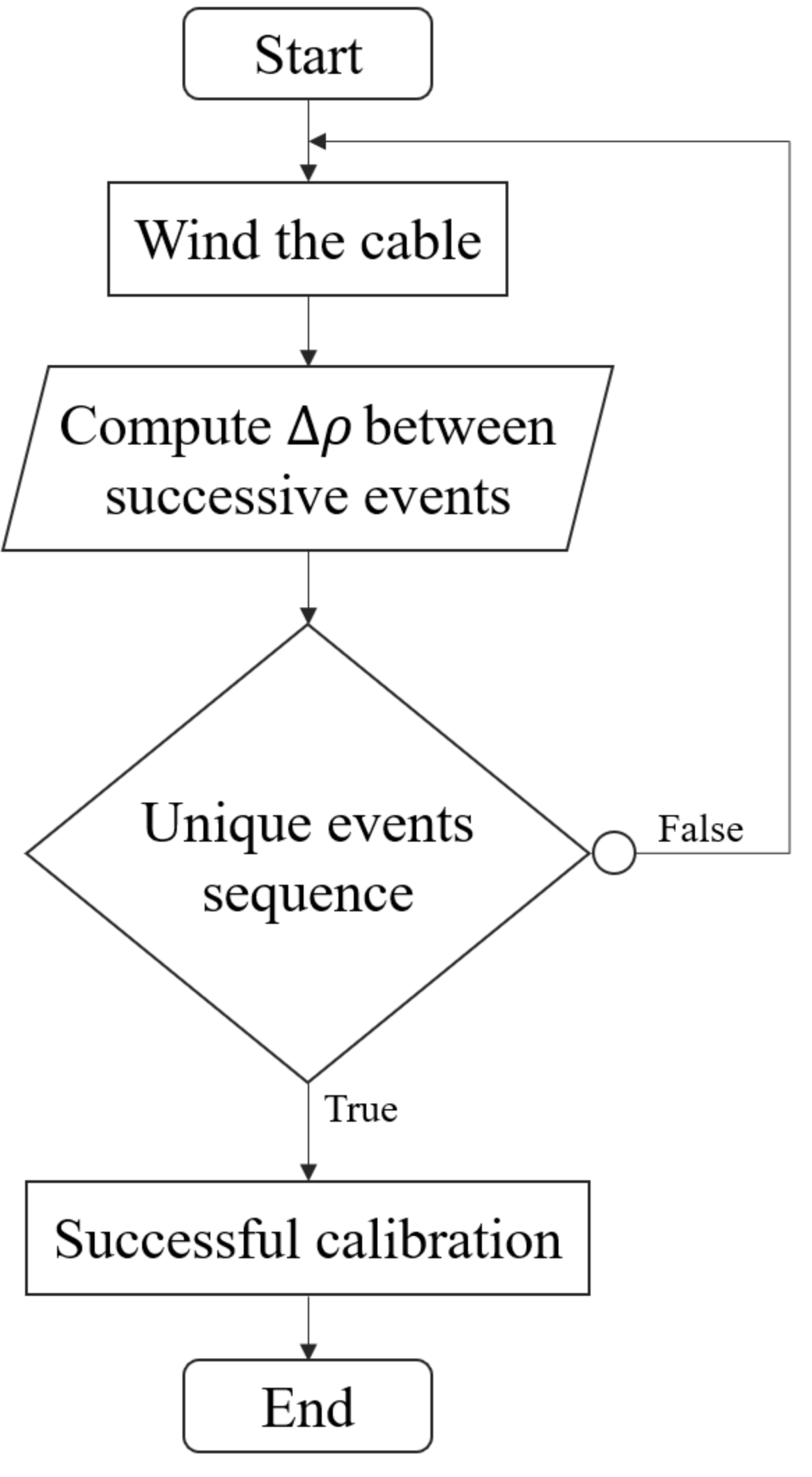}
  \caption{Autocalibration flowchart}
  \label{fig:f5}
\end{figure}

At start, we will wind cable until  first event at $t_0 = t_{i,j}$. This first detection does not allow us to determine which pair $(M_i, S_j)$ is it. We will, therefore, continue to wind cable, until detection of the second mark at $t_{k,l}$. The computing of difference $\Delta\rho$ between successive events will reduce the number of possible cases. If we succeed in identifying sequence of events, then calibration is successful. Otherwise, we continue to wind cable until identification of \emph{events sequence} (Figure\ref{fig:f5}).

\clearpage

%\clearpage
Next detection will limit the number of possible cases. From identification of a unique events sequence, calibration is successful.

\subsection{Autocalibration example}
We will interest in a configuration close to what we may meet in manufacturing facilities. Let consider a CDPR of height $h=3m$, covers a paralleliped $3\times7\times10 m$, with $\rho_{max}=13m$. The distance between successive marks will be variable: \\ $\forall i \in [|1,n_m|], d_i \in \{0.25, 0.50, 0.75, 1.00, 1.25, 1.50, 1.75\} [m]$. The mean value $\bar d = 1m$. Then $d_0 = 0.25 m$, first mark and $n^{th}$ sensor will be placed at a distance of $0.25 m$ from point A. First sensor will be placed on support, by equation (\ref{eq:eqos1}), at:
\begin{center}
\fbox{$\mathbf{\|OS_1\|}= 1m$}
\end{center}
We \emph{boost} cable length by $b = 1m$, by equation (\ref{eq:eqdn}), we get:
\begin{center}
\fbox{$d_n= 3m$}
\end{center}
The distance between successive sensors will be variable:\\
$\forall j \in [|1,n_s|], z_j \in \{0.5, 1.25\} [m]$ \\
The mean value $\bar z$, by equation (\ref{eq:eqmeanz}):
\begin{center}
\fbox{$\bar{z}= 0.875m$}
\end{center}
For the number of sensors, using equation (\ref{eq:eqzn}), $n_s=3$, they will be placed, vertically on the support, as follow:

%\clearpage

\begin{table}[!h]
  \begin{center}
    \caption{Autocalibration example sensors position}
    \medskip
      \label{tab:t1}
    \begin{tabular}{|l|r|}
      \hline
      Sensor &  $\mathbf{\|OS_j\|} \hspace{.1in} [m]$\\
      \hline
      $S_3$  &  2.75\\
      \hline
      $S_2$  &  1.5\\
      \hline
      $S_1$  &  1.00\\
      \hline
    \end{tabular}
  \end{center}
\end{table}

The number of marks, by equation (\ref{eq:eqnm2}):
\begin{center}
\fbox{$n_m= 11$}
\end{center}

They will be placed on cable as follow:
\begin{table}[!h]
  \begin{center}
    \caption{Autocalibration example distance between successive marks}
    \medskip
      \label{tab:t21}
    \begin{tabular}{|l|r|}
      \hline
      Mark &  $\mathbf{\|BM_i\|} \hspace{.1in} [m]$\\
      \hline
      $M_1$  &  12.75\\
      \hline
      $M_2$  &  12.25\\
      \hline
      $M_3$  &  11.50\\
      \hline
      $M_4$  &  10.50\\
      \hline
      $M_5$  &  9.25\\
      \hline
      $M_6$  &  7.75\\
      \hline
      $M_7$  &  6.00\\
      \hline
      $M_8$  &  5.75\\
      \hline
      $M_9$  &  5.25\\
      \hline
      $M_{10}$  &  4.50\\
      \hline
      $M_{11}$  &  3.00\\
      \hline
    \end{tabular}
  \end{center}
\end{table}

Using this configuration, the list of possible events is:
\clearpage

\begin{table}[!h]
  \begin{center}
    \caption{Autocalibration example list of possible events}
    \medskip
      \label{tab:t22}
    \begin{tabular}{|c|c|c|c|c|}
      \hline
      $t_{i,j} \hspace{.1in} [s]$ & $M_i$ & $S_j$ & $\rho_{i,j} \hspace{.1in} [m]$ & $\Delta\rho \hspace{.1in} [m]$\\
      \hline
      0.50 & $M_1$ & $S_3$ & 12.50 & \\
      \hline
      1.00 & $M_2$ & $S_3$ & 12.00 & 0.50\\
      \hline
      1.75 & $M_1$ & $S_2$ & 11.25 & 0.75\\
      \hline
      2.25 & $M_1$ & $S_1$ & 10.75 & 0.50\\
      \hline
      2.75 & $M_4$ & $S_3$ & 10.25 & 0.50\\
      \hline
      $\cdots$ & $\cdots$ & $\cdots$ & $\cdots$ & $\cdots$\\
      \hline
      10.00 & $M_10$ & $S_2$ & 3.00 & 0.25\\
      \hline
      10.25 & $M_{11}$ & $S_3$ & 2.75 & 0.25\\
      \hline
      10.50 & $M_{10}$ & $S_1$ & 2.50 & 0.25\\
      \hline
      11.50 & $M_{11}$ & $S_2$ & 1.50 & 1.00\\
      \hline
      12.00 & $M_{11}$ & $S_1$ & 1.00 & 0.50\\
      \hline
    \end{tabular}
  \end{center}
\end{table}

\begin{figure}[!h]
  \centering
  \includegraphics[width=.7\linewidth]{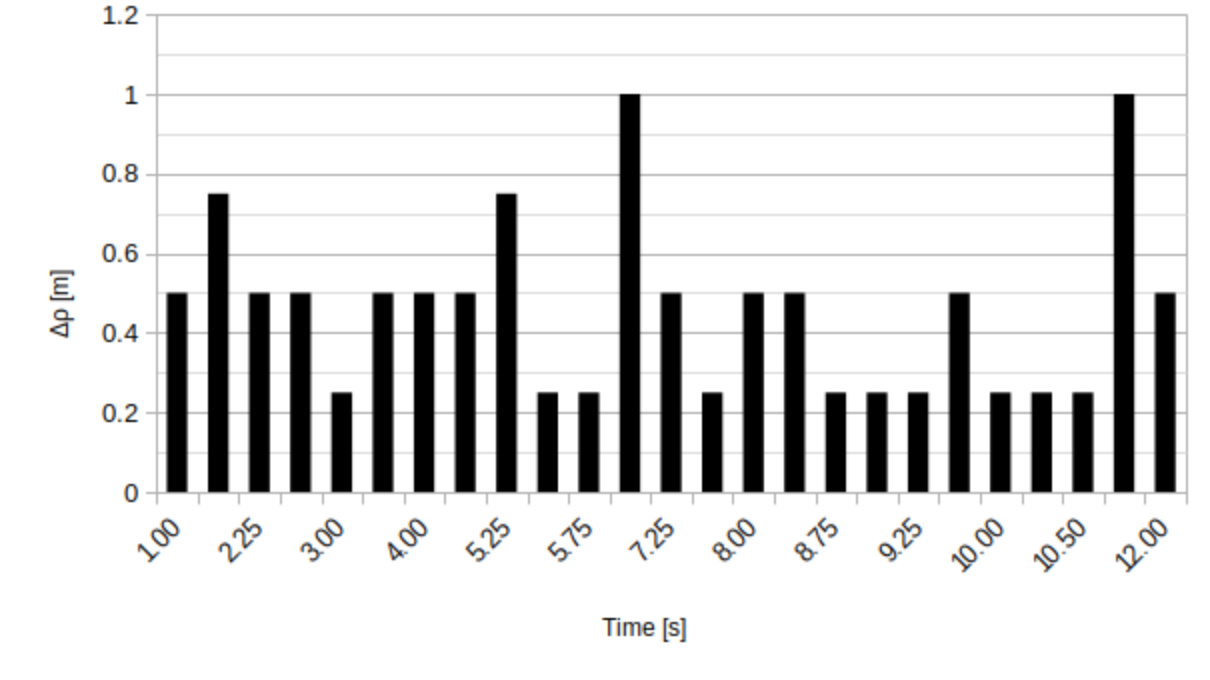}
  \caption{Autocalibration example $\Delta\rho(t)$}
  \label{fig:fae1}
\end{figure}

With respect to the three conditions, total number of events $n_e=26$. $\Delta\rho=f(t)$ is variable (Figure~\ref{fig:fae1}).

\begin{figure}[!h]
  \centering
  \includegraphics[width=.7\linewidth]{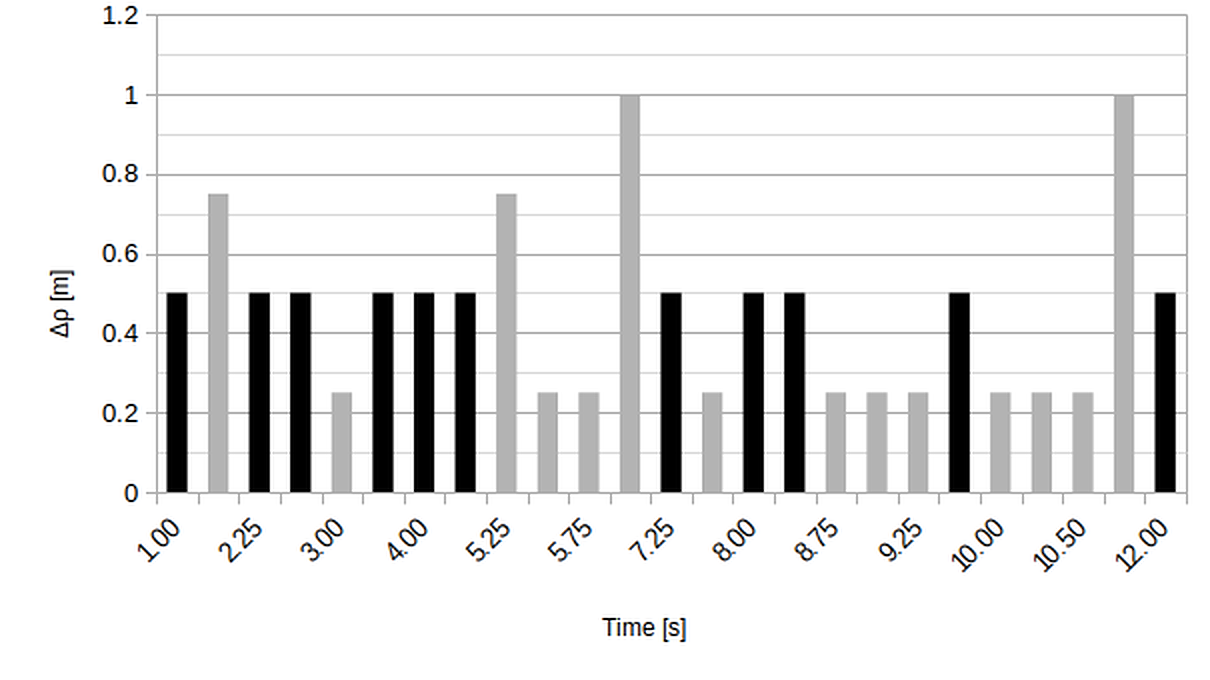}
  \caption{Autocalibration scenario (1)}
  \label{fig:fae2}
\end{figure}

In the beginning, we will suppose that we are somewhere between the first and last events (between $E_{1,3}$ and $E_{11,1}$). We wind the cable, then we get a first detection, our starting position, but this first event does not give us any information about cable length. We, therefore, continue to wind the cable, until the second detection. Then we compute $\Delta\rho$, this is the length we needed for a second detection. Let say $\Delta\rho=0.5m$, which corresponds to several possible cases (Figure~\ref{fig:fae2}).

\begin{figure}[!h]
  \centering
  \includegraphics[width=.7\linewidth]{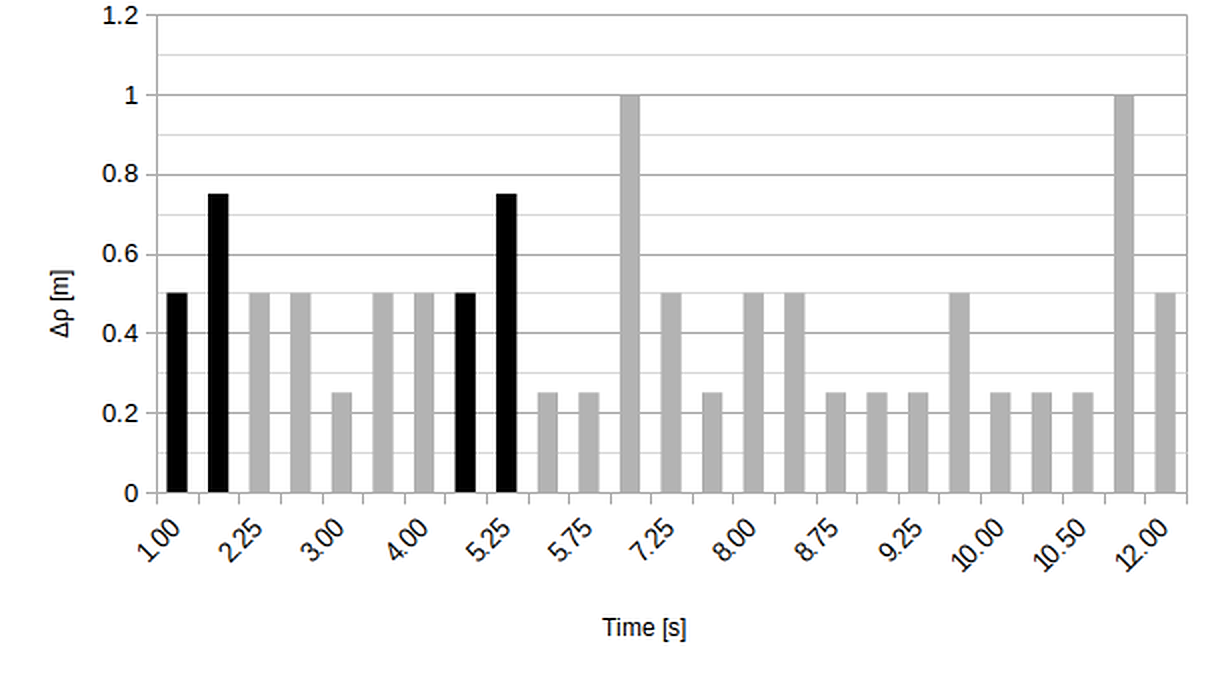}
  \caption{Autocalibration scenario (2)}
  \label{fig:fae3}
\end{figure}

We, therefore, continue to wind the cable, until the third detection. We compute again $\Delta\rho$. Let say, we find $\Delta\rho=0.75m$. We reduce the number of possible cases, bat we still have two cases (Figure~\ref{fig:fae3}).

\begin{figure}[!h]
  \centering
  \includegraphics[width=.7\linewidth]{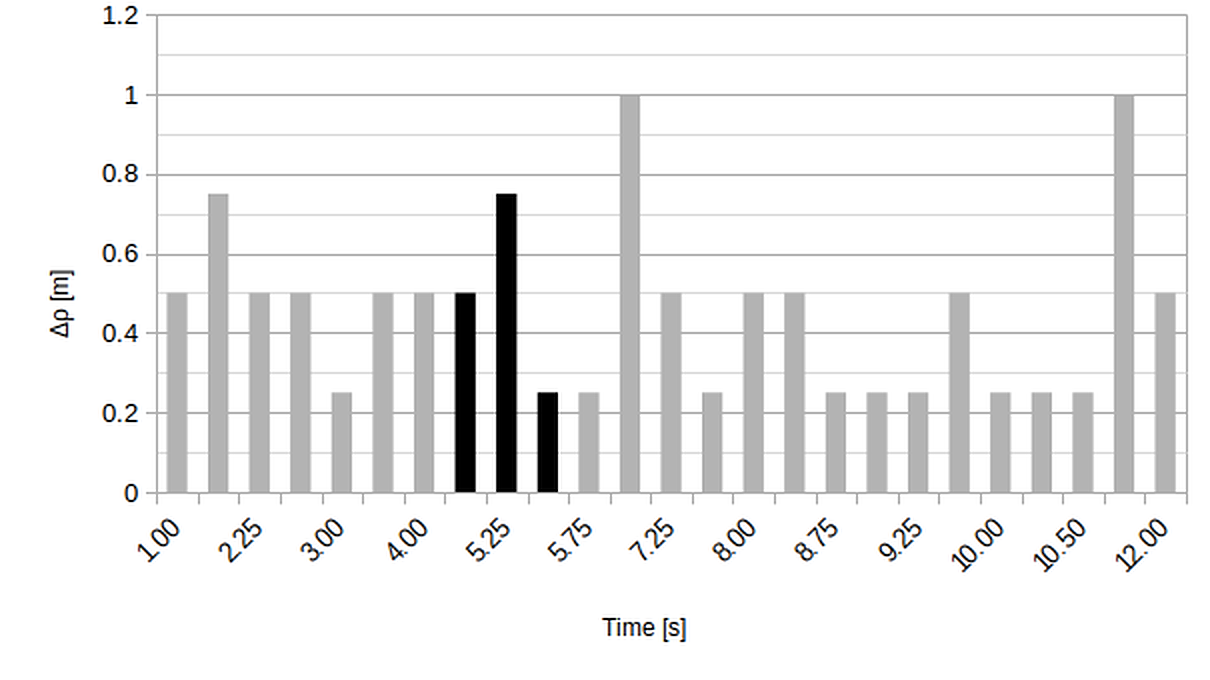}
  \caption{Autocalibration scenario (3)}
  \label{fig:fae4}
\end{figure}

Let's continue to wind the cable, until further detection. We calculate $\Delta\rho$ again. We find $\Delta\rho=0.25m$. So we identify which sequence is it (Figure~\ref{fig:fae4}).

\clearpage

Calibration is done! We get our sequence. The last event corresponds to a cable length $\rho = \mathbf{\|AB\|}=7.50m$. This scenario will be performed automatically by the \emph{controller}.\\

\textbf{Remark 3:} In case where cable is wound over a length greater than $(d_n-d_0)$, without detection; this means that we are on $\overline\mathbf{BM_n}$ segment. In this case cable length will be:
\begin{equation}\label{eq:eqnod}
  \rho=\rho_{n,1}+(d_n-d_0)
\end{equation}

Stroke calibration, by equation (\ref{eq:eqrhoc}), $\rho_c=1m$. Value of length necessary for calibration depends on cable length that is wound in medium $\overline{\Delta\rho}$ and number of distinct events necessary for calibration, this difference depends on standard variation $\sqrt{var(\Delta\rho)}$. To minimize stroke, we must reduce the number of events required for calibration. For this, we will vary $-$ as much as possible $-$ successive distance between marks ($d_i$) and successive distance between sensors ($z_j$). The optimal value corresponds to minimum mean length ($\overline{\Delta\rho}$) and maximum standard variation ($\sqrt{var(\Delta\rho)}$).

%\clearpage

\section{Conclusion}
The Autocalibration Subsystem of the CDPR system fulfills the function \emph{"auto-calibrate cables length"}. It consists of an \emph{instrumented cable}, \emph{inductive sensors} and a \emph{controller} for digital data processing. We have defined seven conditions for subsystem implementation. The number and position of marks and sensors depend on CDPR congestion. Distance between successive marks must be variable, as well as the distance between successive sensors. Metallic marks must be placed on cable, starting from distal point B. Sensors must be placed vertically on support, starting from winch center O.

Each detection of a mark with a sensor corresponds to an event. The number of events depends on the number of marks and one of the sensors; if we increase the number of marks or sensors, then the number of events will increase. For calibration, we analyze the succession of events. Each event corresponds to a cable length. Analysis of the difference in cable length between successive events, reduce the number of possible cases. The identification of event sequence allows cable length calibration. The optimal is to calibrate with least of events, using a minimum of cable length.

\begin{figure}[!h]
  \centering
  \includegraphics[width=.6\linewidth]{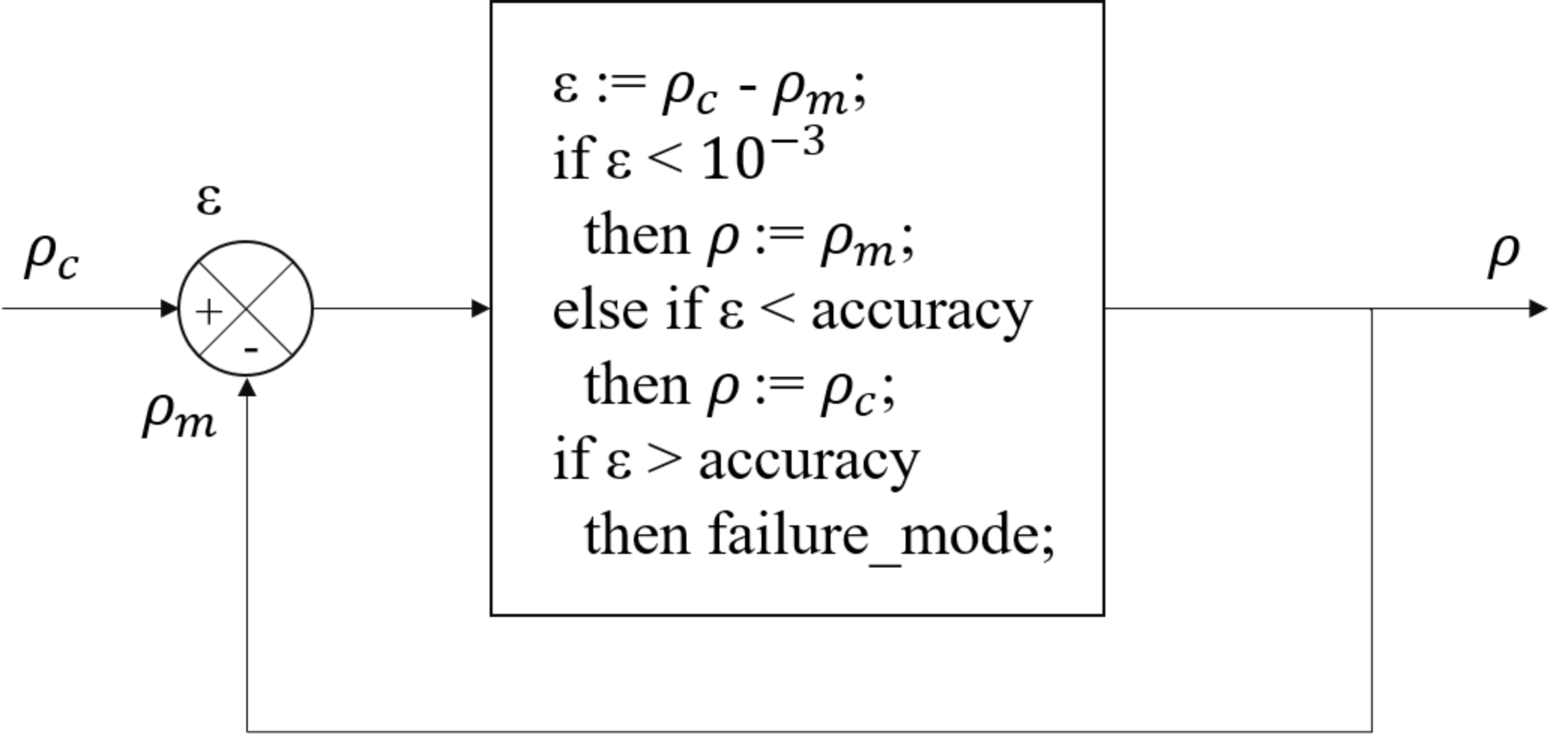}
  \caption{Closed-loop autocalibration approach}
  \label{fig:f6}
\end{figure}

The Autocalibration Subsystem may allow continuously measured cable length adjustment (Figure~\ref{fig:f6}).

Closed-loop control of a parameter consists of linking measurement to a setpoint. Cable length identified by Autocalibration Subsystem corresponds to \emph{setpoint}, that communicated by encoder corresponds to \emph{measurement}. Continuous adjustment of cables length will, therefore, \emph{improve CDPR accuracy}.

%\clearpage

%\section*{Acknowledgement} This work was supported by the French National Research Agency (ANR) under Grant No. ANR-18-CE10-0004 (CRAFT ANR Project).

\bibliographystyle{ieeetr}
\bibliography{references}
\end{document}